\documentclass[sigconf, authordraft]{IEEEtran}

\usepackage{cite}
\usepackage{amsmath,amssymb,amsfonts}
\usepackage{algorithmic}
\usepackage{graphicx}
\usepackage{textcomp}
\usepackage{xcolor}
\usepackage{subfig}
\usepackage{float}
\usepackage{color}
\usepackage{mathrsfs}
\usepackage{caption}
\usepackage{array}
\usepackage{booktabs} 
\usepackage{authblk}
\usepackage{soul}

\begin{document}
\title{Deep Q Learning Driven CT Pancreas Segmentation with Geometry-Aware U-Net}

\author{Yunze Man$^{\dagger}$ \and Yangsibo Huang$^{\dagger}$ \and Junyi Feng \and Xi Li$^*$ \and Fei Wu \thanks{Copyright (c) 2019 IEEE. Personal use of this material is permitted. However, permission to use this material for any other purposes must be obtained from the IEEE by sending a request to pubs-permissions@ieee.org.
\\ \indent Y. Man, Y. Huang, J. Feng, X. Li and F. Wu are with the College of Computer Science and Technology, Zhejiang University, Hangzhou 310027, China (e-mail: {yzman@zju.edu.cn; hazelhuang@zju.edu.cn; fengjunyi@zju.edu.cn; xilizju@zju.edu.cn; wufei@cs.zju.edu.cn})
\\ \indent $\dagger$ Y. Man and Y. Huang contributed equally.
\\ \indent * X. Li \emph{is the corresponding author.}}}

\maketitle

\begin{abstract}
Segmentation of pancreas is important for medical image analysis, yet it faces great challenges of class imbalance, background distractions and non-rigid geometrical features. To address these difficulties, we introduce a Deep Q Network(DQN) driven approach with deformable U-Net to accurately segment the pancreas by explicitly interacting with contextual information and extract anisotropic features from pancreas. The DQN based model learns a context-adaptive localization policy to produce a visually tightened and precise localization bounding box of the pancreas. Furthermore, deformable U-Net captures geometry-aware information of pancreas by learning geometrically deformable filters for feature extraction. Experiments on NIH dataset validate the effectiveness of the proposed framework in pancreas segmentation.
\end{abstract}

\section{Introduction}
\begin{figure}[t]
\setlength{\belowcaptionskip}{-0.5cm}
\centering
\subfloat[0.09\%]{
\begin{minipage}[t]{0.105\textwidth}
  \centering
  \includegraphics[width=1.85cm]{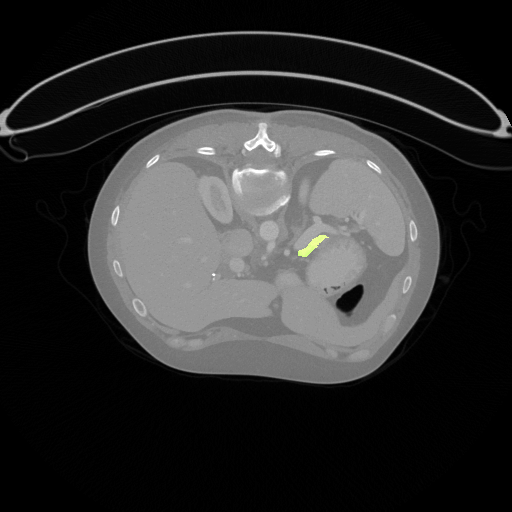}
  \end{minipage}
}
\subfloat[0.76\%]{
\begin{minipage}[t]{0.105\textwidth}
  \centering
  \includegraphics[width=1.85cm]{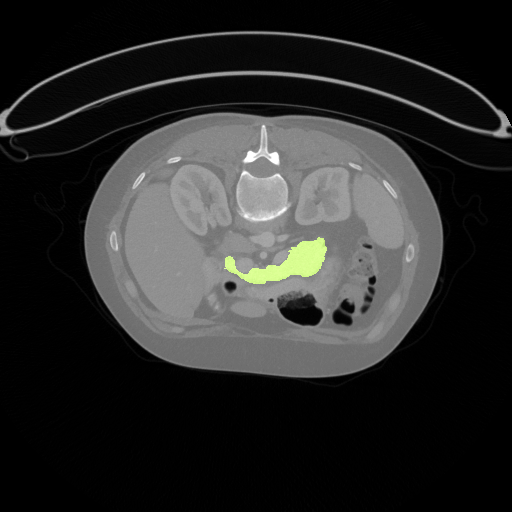}
  \end{minipage}
}
\subfloat[0.43\%]{
\begin{minipage}[t]{0.105\textwidth}
  \centering
  \includegraphics[width=1.85cm]{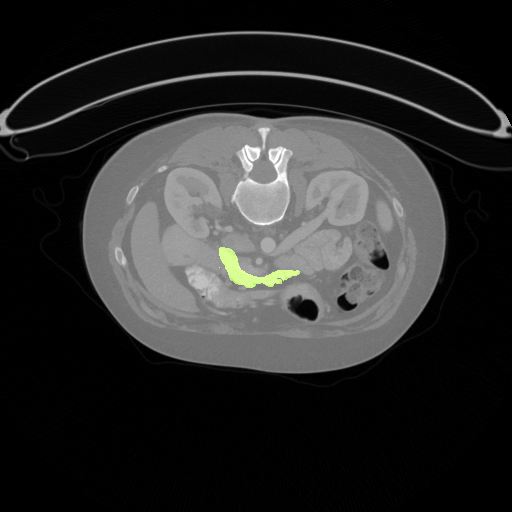}
  \end{minipage}
}
\subfloat[0.19\%]{
\begin{minipage}[t]{0.105\textwidth}
  \centering
  \includegraphics[width=1.85cm]{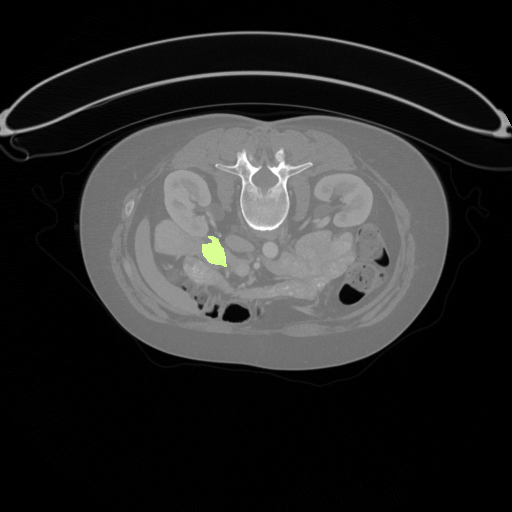}
  \end{minipage}
}
\caption{Illustration of challenges in pancreas segmentation. The images demonstrates the deformation of pancreas and its tininess in size. The pancreas zones (marked as green) vary in geometrical shape and angle. The smallest pancreas region can be less than 0.1\%, and the largest part is no more than 0.8\% of the whole slice. Better viewed in color.}
\label{Challenge_Illustration}
\end{figure}

% The development of deep learning methods for biomedical decision support using medical images is a new trend in healthcare informatics research. Biomarkers extracted from CT, MRI or PET images based on medical image segmentation can serve as phenotype for certain disease and intended for futher radiomics and radiogenomics study.
As an important and challenging problem in medical image analysis,  pancreas segmentation over
a CT volume is typically cast as a voxel-wise classification problem\cite{erdt2011automatic}, which aims to assign
semantic class labels to different CT image regions in a data-driven learning fashion.
Usually, such a learning problem encounters numerous difficulties with small-sample-sized
training, severe class imbalance, and background clutter with confusing distractions.
As shown in Fig ~\ref{Challenge_Illustration},  a pancreas occupies
less than 0.5\% fraction of the entire CT volume, and meanwhile has a visually
blurry inter-class boundary with respect to other tissues. Furthermore,
the pancreas possesses the appearance properties of diverse shapes,
various orientations, and different aspect ratios. Such challenging factors
often makes the quality of data-driven learning degenerate to
extremely biased results (e.g., eroded greatly by non-pancreas regions along
with disrupted segmentation results). Therefore, the focus of this paper is on
setting up an effective data-driven learning scheme for robust pancreas segmentation
with a context-adaptive and environment-interactive pipeline.

Mathematically, the goal of data-driven pancreas segmentation is to build a
mapping function from a 3D CT volume to a 3D segmentation mask. Such a mapping
function is usually formulated as a deep neural network due to its power of
feature representation and discriminative learning~\cite{roth2016spatial,zhu20173d,zhou2017fixed}.
In practice, the pancreas segmentation task is often equipped with
small-sample-sized training images with a tiny portion of pancreas pixels,
which generally leads to inaccurate segmentation results if only using a one-pass
learning strategy. Hence, a two-stage learning framework~\cite{zhu20173d}
is proposed for coarse-to-fine segmentation. In the framework, the coarse segmentation
stage roughly crops out the pancreas regions based on the coarse segmentation results
with a heuristic cropping strategy. Subsequently,
the fine segmentation stage learns another segmentation network to take the coarse localization
results as input and finally output the refined segmentation results.
Therefore, the final segmentation performance relies heavily on
the coarse localization results. However, such a coarse localization process
only adopts a one-pass learning scheme without taking into account the interactions
with its context (encoding the anisotropic geometry properties of the pancreas).
In complicated scenarios, its localization results are usually
unreliable and unstable due to the lack of an effective error correction mechanism.
In addition, the fine segmentation stage is incapable of well extracting the
anisotropic geometry-aware features for adapting to the drastic
pancreas shape deformations, resulting in an imprecise segmentation performance.

Motivated by the above observations, we propose an anisotropic geometry-aware
two-stage deformable deep learning scheme with a contextual interaction mechanism
governed by a deep reinforcement learning (DRL) strategy. The proposed scheme
builds a context-adaptive localization agent that
adaptively learns an environment-interactive
localization error correction policy based on
a deep Q network (DQN). The agent is capable of dynamically and adaptively
adjusting the localization results according to the anisotropic geometry-aware
interaction results with the contextual pancreas-related environment.
Specifically, the DQN is trained by the $\epsilon$-greedy and
replay memory methods to hierarchically zoom into or
shift the candidate region in accordance with a higher intersection-over-union (IoU) ratio
of pancreas.
After the DRL driven pancreas localization, the proposed scheme designs a deformable version
of deep U-Net, which is able to effectively capture the anisotropic geometry-aware
information on the pancreas in face of drastically non-rigid shape deformations.

In summary, the main contributions of this work are summarized as follows:
\begin{itemize}
\item We propose a novel anisotropic geometry-aware pancreas localization scheme based on a context-adaptive
DRL agent, which learns an effective DQN with an adaptive
environment-interactive localization error correction policy. Through the
DQN-guided interactions with the pancreas-related environment, the DRL agent
is able to produce reliable and stable localization results after the policy exploration
process.
\item Based on the DRL-based localization, we further present a deformable deep U-net approach for accurate and
robust pancreas segmentation with  drastically non-rigid shape deformations. The presented approach
has the ability to extract the geometry-aware features for pancreas segmentation, leading to a great performance gain.
To the best of our knowledge, it is innovative to design an anisotropic geometry-aware
deformable deep learning scheme for pancreas segmentation.
\end{itemize}

The rest of this paper is organized as follows. We first review related work in
Sec. 2. The technical details of the proposed anisotropic geometry-aware pancreas segmentation scheme are described in Sec. 3. Sec. 4 presents the experimental results. The paper is finally concluded in Sec. 5.

\section{Related Work}
\begin{figure*}[t]
  \centering
  \includegraphics[width=\textwidth]{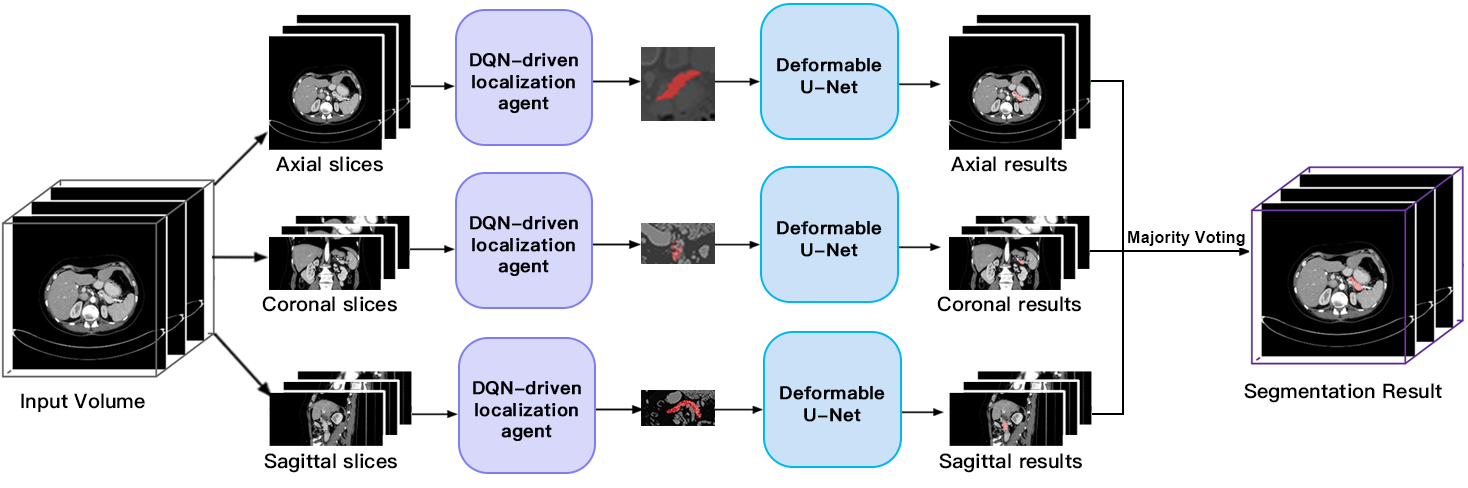}\\
  \caption{
    Overall architecture of our approach. A 3D volume is splitted into 2D slices of three axes. For a single slice, reinforcement learning is first used to crop the region of pancreas, and then deformable U-Net output a mask of the target pancreas. The pixel-wise results of three axes are merged by major voting to formulate the final voxel-wise mask of a 3D volume.
  }
  \label{Overall}
\end{figure*}

\subsection{Pancreas Segmentation} Pancreas Segmentation is one of the hot topics in Medical Image Segmentation, which also includes substructures such as vessel\cite{fu2016retinal} and leision\cite{ValverdeCRGPVRR17,abs-1711-11069}. Deep learning has been widely used in the medical segmentation domain. Gibson et al. \cite{gibson2018automatic} proposes a registration-free deep learning based segmentation algorithm to segment eight organs including pancreas. Li et al. \cite{li2018h} proposes a hybrid densely connected UNet, H-DenseUNet, to segment tumors and livers. Chen et al. \cite{chen2018drinet} present Dense-Res-Inception Net to address the general challenges of medical image segmentation. Pancreas has its own features, in order to deal with great anatomical variability of pancreas, multi-pass structures have been proposed for more accurate segmentation. Roth et al. \cite{roth2015deeporgan} performed a pre-segmented pancreas proposal generating algorithm followed by a proposal refinement convolutional network. This framework is further improved in \cite{roth2016spatial} by segmentation holistically-nested network. Zhou et al. \cite{zhou2017fixed}\cite{yu2018recurrent} designed convolutional network model to localize and segment the pancreas in cyclic manners. In the proposed model, every segmentation stage takes its last segmented zone as its input, and generate a new segmentation map. Zhu et al. \cite{zhu20173d} propose a successive 3D coarse-to-fine segmentation model, consisting of a coarse segmentation network and a fine segmentation network. The coarse-to-fine model utilizes by-pass structure in ResNet \cite{he2016deep} and reaches state-of-the-art mean DSC $84.59\pm4.86\%$ on NIH pancreas dataset \cite{roth2015deeporgan}. However, complexity and distraction of background environment is a typical challenge in pancreas segmentation. Current methods usually have difficulties in well interacting with the contextual environment while localizing the pancreas. Namely, they use pre-defined proposal bounding box or one-pass stage to do the localization. In reality, human vision generally follows a progressive process in localizing pancreas. Inspired by this, we propose a reinforcement Q learning network to model this process.

\subsection{Image Segmentation} Image segmentation has always been a fundamental and widely discussed problem in computer vision \cite{remotesensing16}\cite{texture15}. After Fully convolutional network (FCN) \cite{long2015fully} was proposed, numerous deep convolutional networks have been designed to solve pixel-wise segmentation problems. Badrinarayanan et al. \cite{badrinarayanan2017segnet} and Noh et al. \cite{noh2015learning} presented deep encoder-decoder structures to extract features from input image and generate dense segmentation map from feature maps. Ronneberger et al. \cite{ronneberger2015u} proposed an elegant network, which consists of multiple cross-layer concatenations, to learn from small amount of data, especially in medical image analysis. However, the geometric transformations are assumed to be fixed and known within these networks. To deal with this problem, deformable convolution is brought up in \cite{dai2017deformable} as an alternative to standard convolution, allowing adaptive deformation in scale and shape of receptive field.

\subsection{Object Detection}
In object detection context, region proposals extraction and proposal evaluation are highly correlated. Based on initial Region-based CNN(R-CNN)\cite{RCNN}, advanced approaches identify the high correlations between region generation and evaluation, and introduce Fast R-CNN\cite{Fast} and Faster R-CNN \cite{Faster} to allow more interaction in the mechanism by using a Region Proposal Network(RPN) and allowing shared convolutional computation among regions. Moreover, generic sample learning is usually confronted with overfitting, a problem seldom happens for strategy learning methods. For strategy learning algorithms like Deep Q Network(DQN)\cite{Hierarchy}, the exploration space is large and the ε-greedy policy introduces randomness, which adds to the model’s complexity and makes it hard to overfit. Furthermore, for a sample learning setting, when a model’s complexity is reduced to fit a small dataset, the model may not be powerful enough to learn a good representation and yield promising performance. However, strategy based approaches usually breaks down a global task into a set of subtasks, and the model is trained to perform well at each local stage, which reduces the difficulty in learning and makes it plausible to approach a complex problem with a relatively simple model. Besides, DQN corroborates the interaction between region extraction and evaluation by encoding them into pairs of action and reward in a Markov Decision Process. With a well-qualified exploration space, the DQN is promised to capture the prior about pancreas location even with limited data. Previous work \cite{Active,Hierarchy,Tree} has proposed DQN for efficient object detection, and recent work \cite{DQN_landmark,DQN_breast} applied it to anatomical landmark and breast lesion detection, suggesting its competence in MIA. Nonetheless, the incorporation of DQN into more complex Medical Image Segmentation problems has never been expolred, which is the focus of our work.

\section{Approach}

\subsection{Problem Definition}
Our task is to segment the pancreas out of a given CT scan. The problem setting is given as follows. Given a 3D CT volume $X$ of size $[W,H,D]$, our goal is to learn a function $F$ to output a pixel-wise segmentation map $F(X)$ which is as close to ground truth map $Y$ as possible. The overall function $F$ consists of two modules, a localization process $Loc$ followed by a segmentation process $Seg$. Through the localization process, the model $Loc(X)$ will output region of interest(ROI) which is zoomed into pancreas area. Then $Loc(X)$ is fed to segmentation pipeline to generate the final map. The overall framework can be summarized as:

\begin{equation}\label{overall_eq}
F(X) = Seg(Loc(X))
\end{equation}

In order to make full use of 3D volume information, segmentation is done in three directions separately. As shown in Fig.2, the original 3D input $X$ is sliced to be $X_s$, $X_c$ and $X_a$, corresponding to sagittal, coronal and axial views. For each view, we train a view-specific network. These three networks are of the same structure, but are trained separately, each with data of one single view. We do so out of the fact that different view of the 3D volumn have very different visual features and image size. Each slice goes through $Loc$ and $Seg$ processes and obtain its segmentation result, both $Loc$ and $Seg$ networks are able to accept unfixed image size. All slices of the same view will be re-piled together after segmentation to generate a 3D segmented volumn. With the three segmented volumns, $Seg(Loc(X_s))$, $Seg(Loc(X_c))$ and $Seg(Loc(X_a))$, majority voting is performed to refine the segmentation result and get the final mask. The detailed localization and segmentation processes on each slice are given in the following sub-sections.

\subsection{DRL-driven localization} 

In our work, the goal of \emph{Loc} phase is to obtain a bounding box for the target pancreas. Considering the high variability of pancreas morphology and location, we develop a geometry-aware pancreas localization scheme driven by context-adaptive Deep Reinforcement Learning(DRL). The aim of our localization \emph{agent} is to shrink to a tightened bounding box of pancreas step by step by interacting with given CT slices (see Fig.\ref{DQN}). This localization procedure is discrete and sequential, thus is suitable to be modeled as a Markov Decision Process(MDP). With the goal of maximizing the reward $r$ that reflects the localization accuracy, the \emph{agent} successively transforms from a current state $s$ to a new state $s'$ by performing one of the pre-defined actions $a \in A$. The action set $A$, state $s$ and reward $r$ of our proposed MDP will be parameterized afterwards. Our DRL-driven agent self-learns the localization strategy by constantly updating a Q-network(see Fig.\ref{DQN}) to fit the value function $Q(s,a)$ for each state-action pair.

We hereby first define how the MDP is parameterized in our localization scenario, and then how Deep Q Learning is implemented.

\begin{figure}[t]
\setlength{\belowcaptionskip}{-0.5cm}
  \centering
  \includegraphics[width=0.5\textwidth]{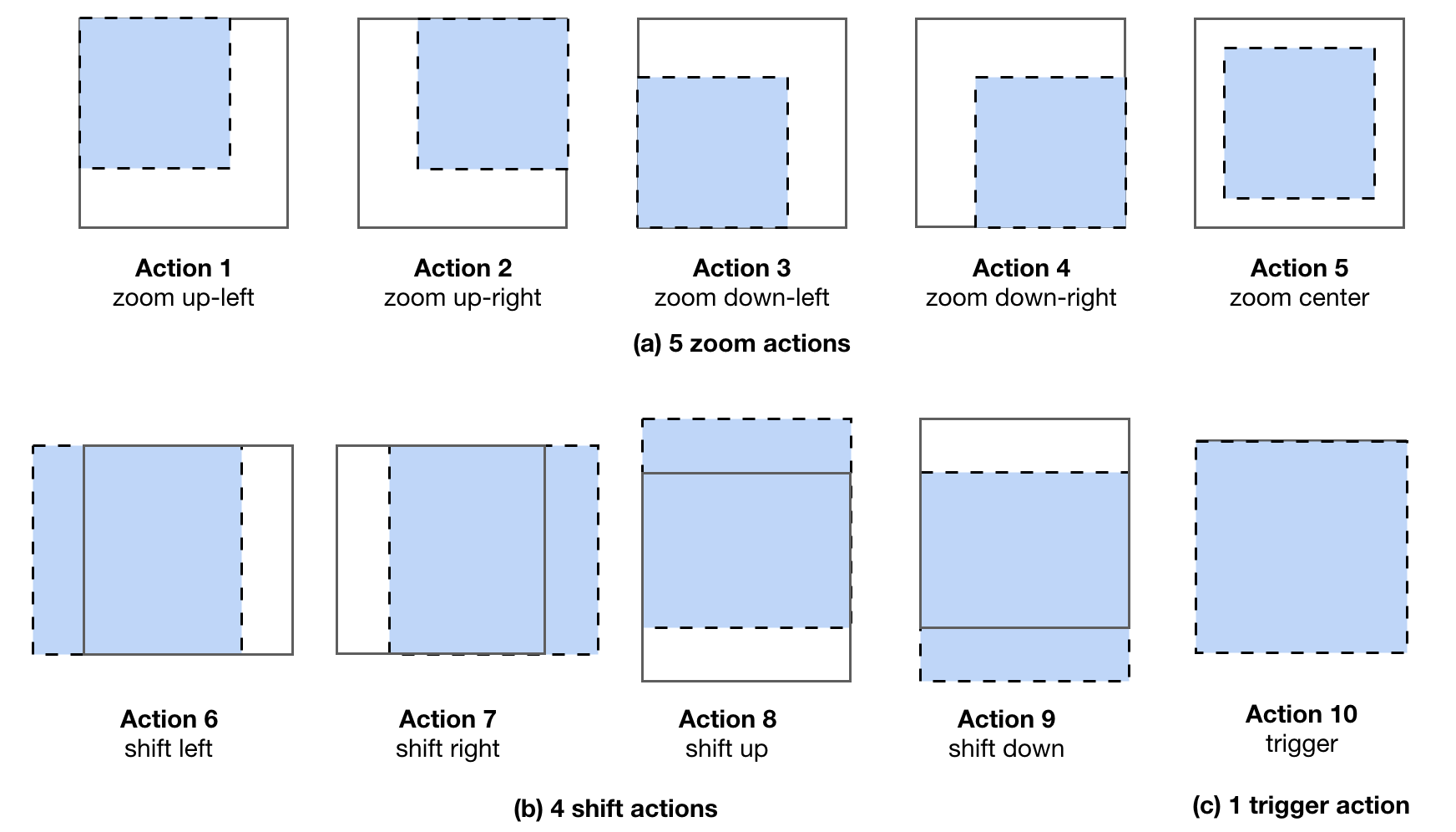}\\
  \caption{
    Illustration of (a)zoom, (b)shift and (c)trigger actions. The original region is denoted by solid line with white filling, and the new region is denoted by dotted line with blue filling. Better viewed in color.
  }
  \label{action}
\end{figure}

\begin{figure}[t]
  \centering
  \includegraphics[width=0.45\textwidth]{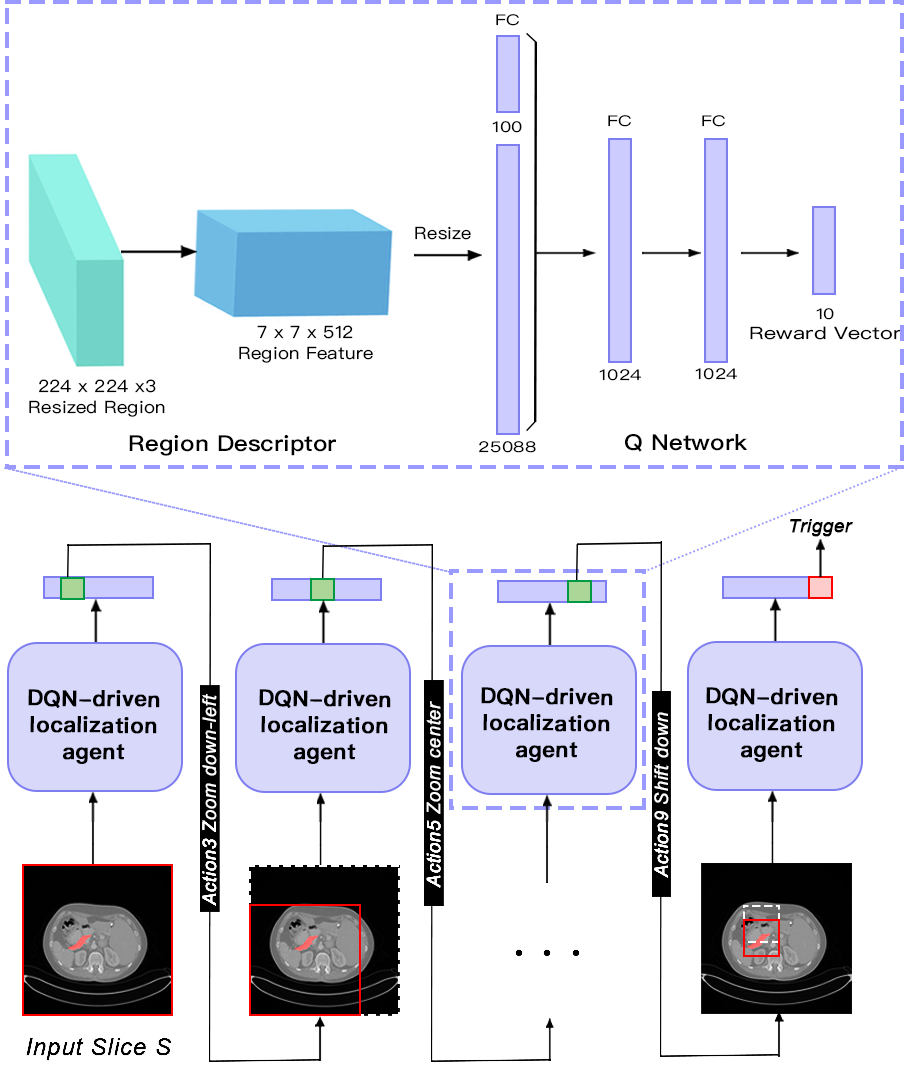}\\
  \caption{
     Illustration of DRL-driven localization. For an input slice S, a sequence of actions are taken to select new regions to shift attention to. Target pancreas is annotated in red. The red solid box indicates the new region, while grey dashed box indicates the old one. The agent is implemented with a Q-network, whose detailed architecture is shown in the dashed box at top. Better viewed in color.
  }
  \label{DQN}
\end{figure}

\noindent
\textbf{MDP formulation}  As illustrated in Fig.\ref{action}, there are three types of \textbf{actions}: 
\begin{itemize}
  \item Five Zoom actions which select a subregion to shift attention to,
  \item Four Shift actions which select a neighbor region to shift attention to,
  \item One Trigger action which indicates that the localization is accurate enough or the search is ended because of maximum steps of action.
\end{itemize}

Specifically, we have five zoom actions (see Fig.\ref{action}(a)) to shrink both side lengths of the region to its 3/4 towards different directions. They are designed to facilitate the search of pancreas in various region scales. Four shift actions(see Fig.\ref{action}(b)), on the other hand, move the window by 1/4 of the current window size horizontally or vertically, helping to refine the current region and perform changes of focus. The next region is then deterministically obtained after taking one of the nine moving actions above. Our agent will terminate in two conditions, either when the trigger action is selected, or when the agent reaches the maximum search step. A trade off is made on the selection of maximum search number --- a larger value introduces larger searching space, while can possibly result in a smaller bounding box in the meanwhile. It is worth mentioning that the combination of zoom and shift actions in opposite directions (e.g. zoom down-right and shift left) plays an important role in precise localization, which will be examined in Sec. 4.

The \textbf{state} of our agent contains two types of information, namely where is the agent's attention, and how does the agent shift to its current attention window. A region descriptor of the current region answers the `Where' question, and a memory vector which replays recent actions taken by the agent indicates `How'. As stated in \cite{AlphaGO}, a memory vector that encodes the state of this refinement procedure is useful to stabilize the search trajectories. The memory vector captures the last 10 actions that the agent has already performed, and the number is designed to guarantee that the agent could zoom closely enough in search for the pancreas.

The \textbf{reward} function $r(s,a)$ reflects the localization accuracy of pancreas when taking action $a$ under state $s$, and is classified as $r_m(s,a)$ for moving actions(zoom and shift), and $r_t(s,a)$ for a trigger action. We adopt the simple yet indicative localization quality measurement, Intersection-over-Union (IoU) between the current window $w$ and the ground-truth pancreas mask $g$, which is defined as $IoU(w,g) = area(w \cap g)/area(w \cup g)$. 
It is worth mentioning that we apply modified IoU here, in that we use the ground-truth mask instead of the ground-truth bounding box of the pancreas, because it better preserves geometry information. The modified IoU reflects the trade-off between window size and pancreas retrieval. A large bounding box with the entire pancreas retrieved, and a tight bounding box with most of the pancreas lost will both render poor IoU.

For nine moving actions, we request improvement in IoU in the definition of reward function. With a movement from state $s$ to state $s'$ after taking action $a$, the window transforms from $w$ to $w'$, and corresponding reward is formulated as:

\begin{equation}
\label{reward-m}
  r_m(s,a) = sign(IoU(w',g) -IoU(w,g))
\end{equation}

As Equation \ref{reward-m} indicates, this binary reward function returns +1 or -1, which reflects more clearly which actions can drive the agent towards the ground-truth. A reward is given to those actions which transform into a windows $w'$ with a greater IoU. Otherwise, the actions are penalized. 

For the trigger action, we pay extra attention to the recall of pancreas. With ground-truth mask $g$ and window $w$, the $Recall(w,g)$ is defined as below.

\begin{equation}
\label{Recall}
Recall(w,g) = area(w \cap g)/area(g)
\end{equation}
A trigger reward shown in Equation \ref{reward-t} is positive if both the IoU and the Recall rate of current window $w$ and ground truth $g$ are both greater than their thresholds $\tau_{IoU}$ and $\tau_{Recall}$, and negative otherwise. 

\begin{small}
\begin{equation}
\label{reward-t}
r_t(s,a)=
  \left\{  
             \begin{array}{lr} 
             +\ \sigma , & \ Recall(w,g) > \tau_{Recall}, \  IoU(w,g) > \tau_{IoU} \\
             -\ \sigma , & \ \text{Otherwise\ \ \ \ \ \ \ \ \ \ \ \ \ \ \ \ \ \ \ \ \ \ \ }
             \end{array}  
\right.  
\end{equation}
\end{small}

$\tau_{IoU}$ is designed to locate the target pancreas. $\tau_{Recall}$ guarantees the retrieval of the target's main part.

\textbf{Deep Q-learning}  The optimal policy of maximizing the overall reward of running an episode starting from the whole slice is learned with reinforcement learning. With the MDP given above, we resort to the Q-learning algorithm to iteratively updates the action-selection policy using a Bellman equation. In state $s$, the Q-value of a specific action $a$ is decided by a Bellman equation $Q(s,a)$ given below:

\begin{equation}
\label{loss}
Q(s,a) = r + \gamma max_{a'}Q(s',a')  
\end{equation}
where $r$ is the immediate reward and $max_{a'}Q(s',a')$ represents the best future reward with action $a'$ at state $s'$. $\gamma \in (0,1)$ is the discount factor, and drives the agent to pursue both the immediate and long-term rewards.

The deep Q-network we use is proposed by \cite{AlphaGO}, and detailed architecture is illustrated in Fig. \ref{DQN}. When a region is selected, its feature descriptor is connected with a history vector to constitute a state, and the state vector is then fed into a multi-layer perceptron which predicts the Q-value of 10 actions with current state. The action with best Q-value is selected to crop new region from the original slice, forming a computational loop for pancreas localization. The update for the network weights at the $i^{th}$ iteration $\theta_{i}$ with transition ($s, a, r, s'$) is:

\begin{small}
\begin{equation*}
\label{bellman}
\theta_{i+1} = \theta_{i}+\alpha(r + \gamma max_{a'}Q(s',a';\theta_{i+1})-Q(s,a;\theta_{i}))\nabla_{\theta_{i}}Q(s,a;\theta_{i})
\end{equation*}
\end{small}
where $\alpha$ is the learning rate.

Two techniques are adopted to achieve more efficient and stable training. First, always using the predicted action by Q-network may mislead the agent, thus a balance between exploration and exploitation is required and $\epsilon$-greedy is employed. During training, the agent can select a random action with probability $\epsilon$, or select a predicted action generated by the Q-network with probability 1-$\epsilon$. Also, strong correlation between consecutive experiences may lead to oscillation, hence a replay memory following \cite{AlphaGO} is used to store the experiences of the past episodes, which allows one transition to be used in multiple updates and breaks the short-time correlations between training samples. A mini batch is randomly selected from replay memory when Q-network is updated.

\subsection{Deformable Segmentation}
\textbf{Network Architecture} For $Seg$ phase, we use U-Net with deformable convolution to extract the features and segment the pancreas. U-Net has been widely used and proved useful in biomedical image processing. As illustrated in Fig.5, the network is composed of an encoder path and a decoder path, both consists of 4 units. Each unit in encoder path consists of two convolutional layers followed by a downsampling layer, and each unit in decoder path consists of a deconvolutional layer (upsampling layer) followed by two convolutional layers. In order to extract the non-rigid features of pancreas, we replace the standard convolution with deformable convolution in encoder phase.

Given the localization result $Loc(X)$, the segmentation phase will then output a $Seg(Loc(X))$, the pixel-wise segmentation map. As this is a two-class segmentation task with class imbalance, we follow \cite{milletari2016v} and use Dice similarity coefficient(DSC) rather than cross entropy to measure the similarity. Given prediction map $F(X)$ and ground truth $Y$, the Dice loss is formulized as

\begin{figure}[!ht]
  \centering
  \setlength{\belowcaptionskip}{-0.5cm}
  \includegraphics[width=0.49\textwidth]{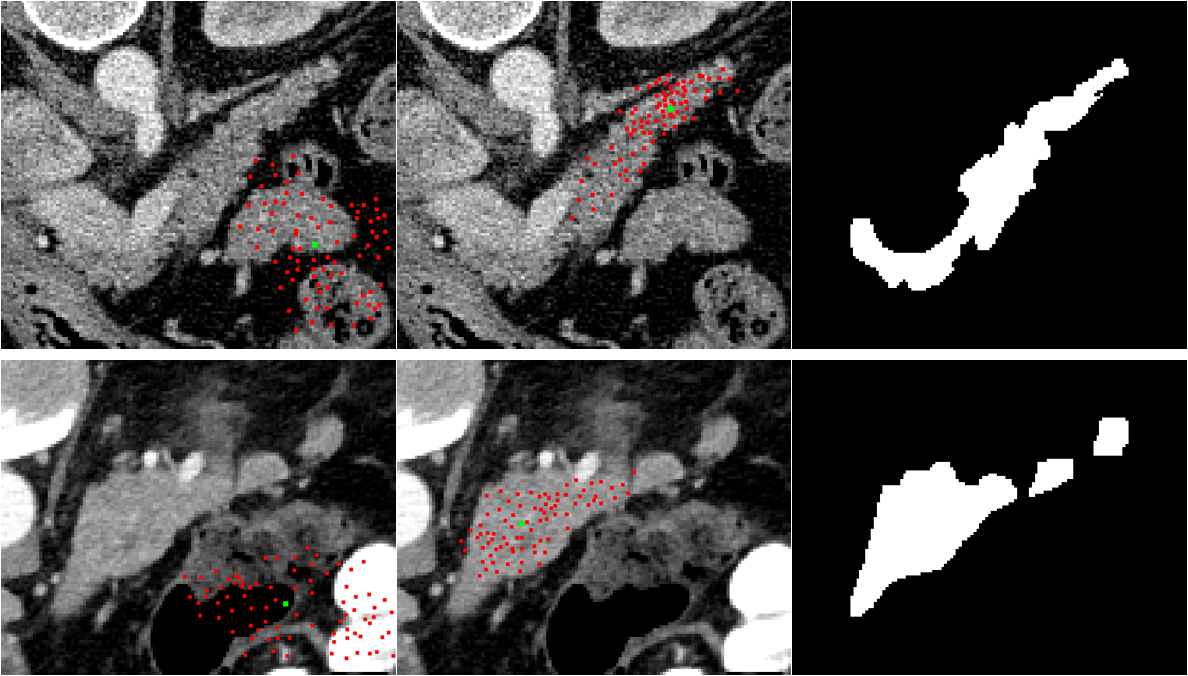}\\
  \vspace{0.2cm}
  \caption{Visualization of our deformable kernel. Each row is a test case in the segmentation phase. The images from left to right are offset kernel for background activation point, offset kernel for pancreas activation point and ground truth mask, respectively. {\color{green}Green dots} are activation points, and {\color{red}red dots} represent the receptive field of that activation point.}
\label{Deformable}
\end{figure}

\begin{small}
\begin{equation}\label{Dice_loss}
DL(F(X), Y) = - DSC(F(X), Y) = -{\frac{2\times \mid{F(X)\cap Y}\mid}{\mid F(X)\mid + \mid Y\mid}}
\end{equation}
\end{small}

\textbf{Deformable Convolution} As discussed in previous sections, segmenting organs is challenging because of their special 3D geometrical shape. For example, a pancreas has a head end and a tail end, and these two ends have very different size and shape. It is hard to extract good features from both ends with receptive field of the same size. What's more, pancreas, as an organ, is non-rigid, which means its shape and size vary in a very wide range. Unlike objects such as cars, roads or planes whose geometrical features maintain a relative rigid state, pancreas can be observed in any perspective in different slices within the same volume. In order to address the above challenges, we use deformable convolution to replace the standard convolution.

\begin{figure*}[!t]
\setlength{\belowcaptionskip}{-0.5cm}
  \centering
  \includegraphics[width=\textwidth]{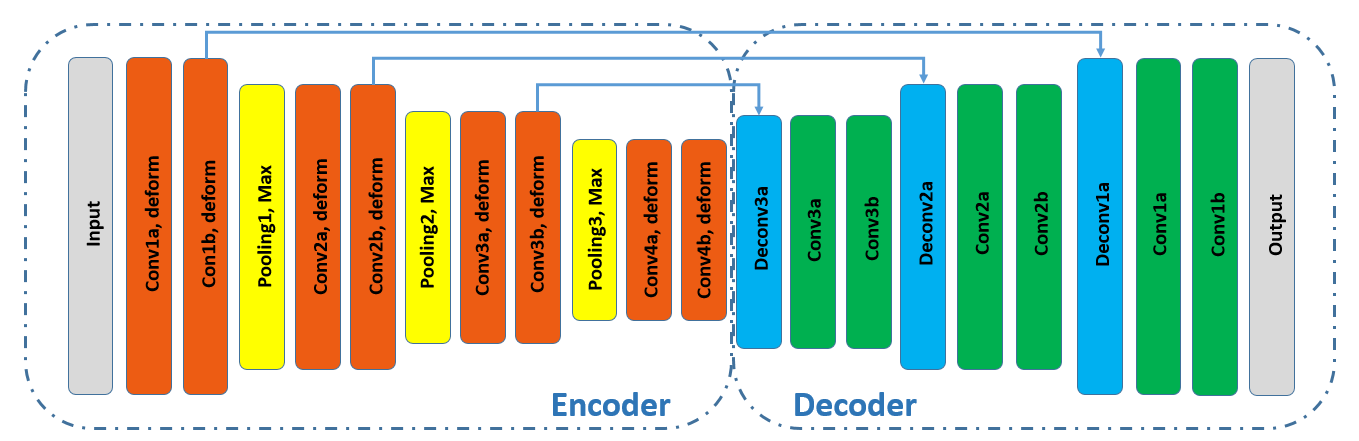}\\
  \caption{
    Illustration of the deformable U-Net architecture (best viewed in color). Orange rectangles stand for deformable convolution layers, and green ones stand for standard convolution layers. Each convolution and deconvolution layer are followed by a ReLU activation, which is omitted in the graph. The output layer is a $1 \times 1$ layer followed by a sigmoid to give a binary classification on each pixel.
  }
\end{figure*}

Deformable convolution is a mechanism which allows receptive field to be learned rather than be fixed in standard convolution. It adds offsets to the kernel location in standard convolution by adding additional convolution layers after each feature map. By introducing free form deformation of the sampling grid, different parts of objects can have different and learnable receptive field. This internal mechanism is consistent with the nature of organs, so it can extract better features from pancreas image.

For a standard convolution, given input feature map $X'$, each pixel $p_0$ in the output feature map $Y$ is given by

\begin{equation}\label{stand_conv}
Y(p_0)=\sum_{p_n\in R}{w(p_n)\cdot X'(p_0+p_n)}
\end{equation}
where $p_n\in {(-1 -1), (-1, 0), ..., (0, 1), (1, 1)}$ for a 3 $\times$ 3 kernel, and $w(p_n)$ denotes the weight of kernel on $p_n$ position.

For a deformable convolution layer, additional offsets are introduced and Equation \ref{stand_conv} becomes

\begin{equation}\label{deform_conv}
Y(p_0)=\sum_{p_n\in R}{w(p_n)\cdot X'(p_0+p_n+\Delta p_n)}
\end{equation}
where $\{\Delta p_n \in n=1,\cdots,N\}$ are learnable offsets of convolution kernel. Note that $\Delta p_n$ is not necessarily integer, so the value $X'(p_0+p_n+\Delta p_n)$ is calculated by bilinear interpolation.

Generally speaking, the offset map is obtained by adding a convolutional layer after the last feature map. For every pixel in the feature map, two offsets are related to it, each for one axis. After shifting the coordinate of convolutional kernel, the feature is extracted by performing the convolution operation. The deformable kernel is exemplified in Fig.\ref{Deformable}.

\subsection{Inference Procedure}
Given a test image $X^{test}$, it is sliced in three directions and get $X^{test}_s$, $X^{test}_c$ and $X^{test}_a$. Each slice will go through localization and segmentation phase and forms the segmentation map, for example $Y_s = Seg(Loc(X^{test}_s))$ is the prediction map on sagittal view. The value in segmentation map is then binarized, value bigger than 0.5 will be classified as pancreas and set to 1, otherwise it is classified as background and set to 0. After generating the segmentation map on 3 different views $Y_s$, $Y_c$ and $Y_a$, we use majority voting to get an accurate final segmentation map, which is given by

\begin{equation}\label{Major_voting}
Y = Majority(Y_s, Y_c, Y_a) = \biggl\lfloor{\frac12 + \frac{Y_s + Y_c + Y_a}{3}}\biggr\rfloor
\end{equation}
where all matrix operators are element-wise. The majority voting makes use of 3 directions' information by making sure that a pixel is finally classified as pancreas only when at least two of the views take it as pancreas.

\section{Experiments}

In this section, we evaluate the performance of pancreas localization (see Sec. 4.3) and segementation (see Sec. 4.4) by conducting extensive evaluations on a benchmark dataset (see Sec. 4.1). Furthermore, we perform an in-depth ablation study (see Sec. 4.5) to evaluate the performance of our method. Additional details about our network architecture and training procedure are reported in Sec. 4.2.

\subsection{Dataset} 

Following previous work of pancreas segmentation\cite{roth2016spatial,dou20163d,zhou2017fixed,zhu20173d}, we evaluate our approach on NIH pancreas segmentation dataset\cite{roth2015deeporgan}, which is the largest and most authoritative Open Source Dataset in pancreas segmentation. The dataset contains $82$ contrast-enhanced abdominal CT volumes. The resolution of each CT scan is $512 \times 512 \times L$, where $L \in [181, 466]$ is the number of sampling slices along the long axis of the body. The slice thickness varies from $0.5mm$ to $1.0mm$. Only the CT slices containing the pancreas are used as input to the system. Following the standard cross-validation strategy, we split the dataset into 4 fixed folds, each of which contains approximately the same number of samples. We apply cross validation, i.e., training the model on 3 out of 4 subsets and testing it on the remaining one. We measure the segmentation accuracy by computing the Dice-Sørensen Coefficient (DSC) for each sample. This is a similarity metric between the prediction voxel set Z and the ground-truth set Y, with the mathematical form of $DSC(z, y) = \frac{2\times \mid{z\cap y}\mid}{\mid z\mid + \mid y\mid}$. For DSC evaluation, we measure the average with standard deviation, max and min statistics over all testing cases.

\begin{figure}[ht]
  \centering
  \setlength{\belowcaptionskip}{-0.5cm}
  \includegraphics[width=0.45\textwidth]{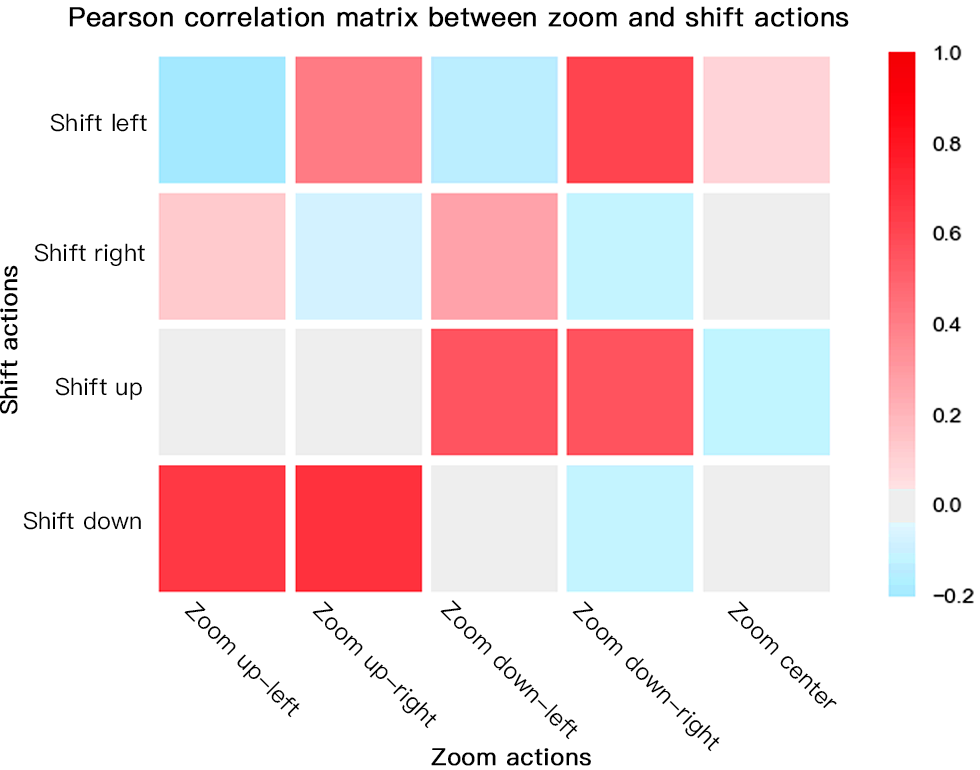}
  \caption{Pearson correlation matrix between zoom and shift actions in axial localization sequences. Significant correlations ($>$0.5) are detected for action pairs of opposite directions, including shift left and zoom down-right, shift up and zoom down-left, shift down and zoom up-right, etc. Better viewed in color.}
  \label{freq}
\end{figure}

\begin{figure*}[t]
  \centering
  \includegraphics[width=\textwidth]{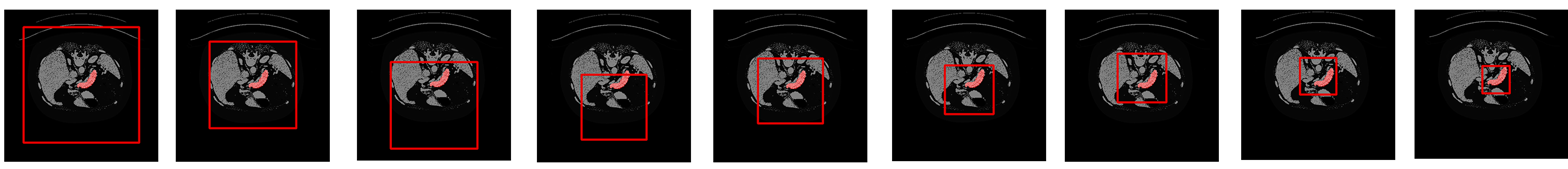}\\
  \caption{
    Sample localization sequence of an axial slice. The pancreas is annotated in red, and the intermediate results of localization are given with a red box. A result bounding box which is visually tightened and precise is generated after taking 9 actions. Although in step 4, our localization window loses part of the pancreas by taking a zoom action, a shift up in step 5 is immediately adopted to make up for the loss. These two steps reflect the error correction capabilities of our model. Better viewed in color.
  }
  \label{sequence}
\end{figure*}

\begin{figure}[t]
  \centering

  \includegraphics[width=0.45\textwidth]{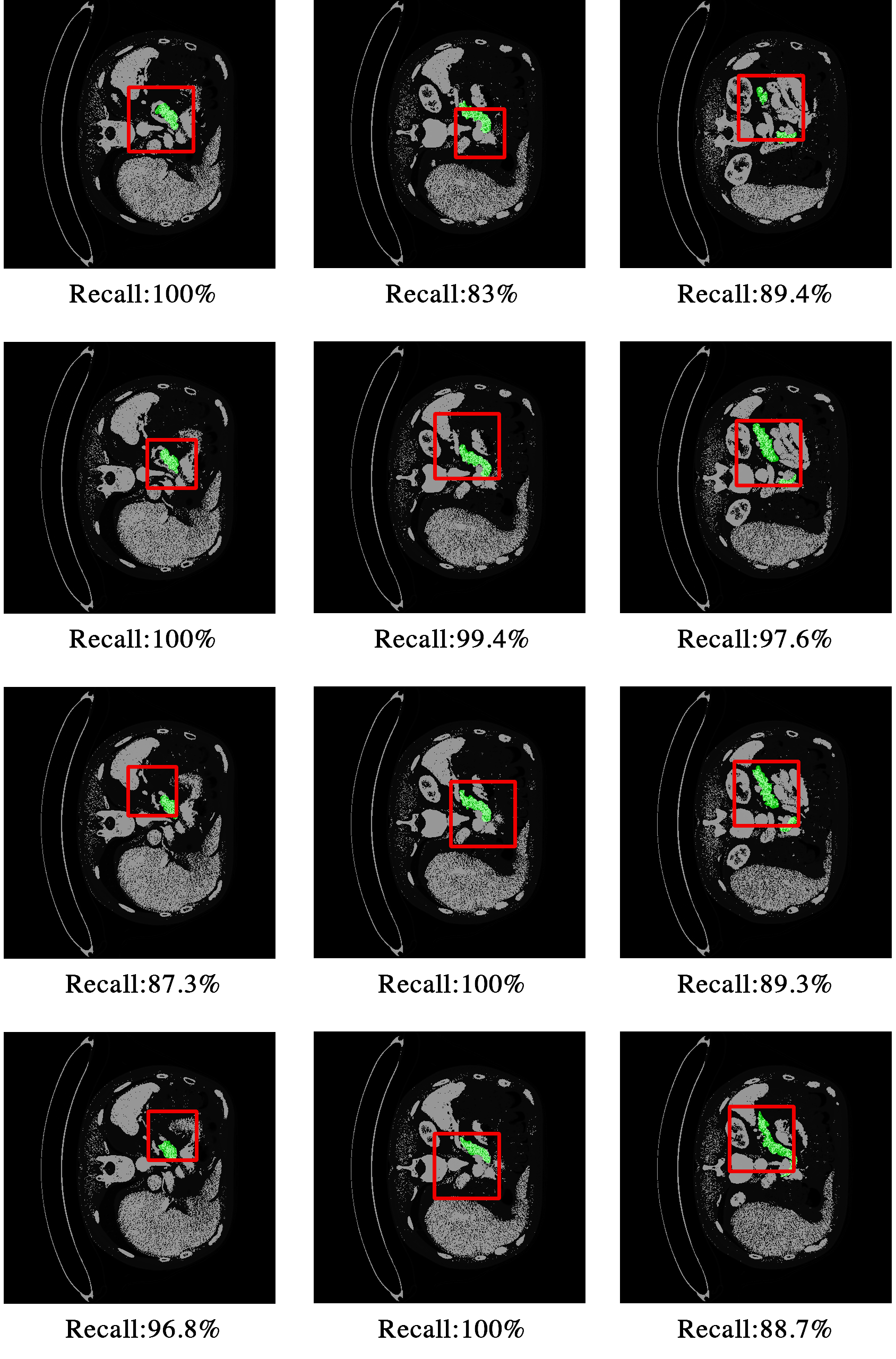}\\
  \setlength{\belowcaptionskip}{-0.2cm}
  \caption{
    Sample localization results of axial slices in volume 76. The pancreas is annotated in green, and the localization result is shown with a red bounding box. Small (pancreas fraction $< 0.1$, shown in first column), medium (pancreas fraction $\in (0.1,0.15)$), shown in second column) and large (pancreas fraction $> 0.15$, shown in third column) pancreas all achieve good localization results. Better viewed in color.
  }
  \label{volume76}
\end{figure}

\begin{figure}[t]
  \centering
  \includegraphics[width=0.45\textwidth]{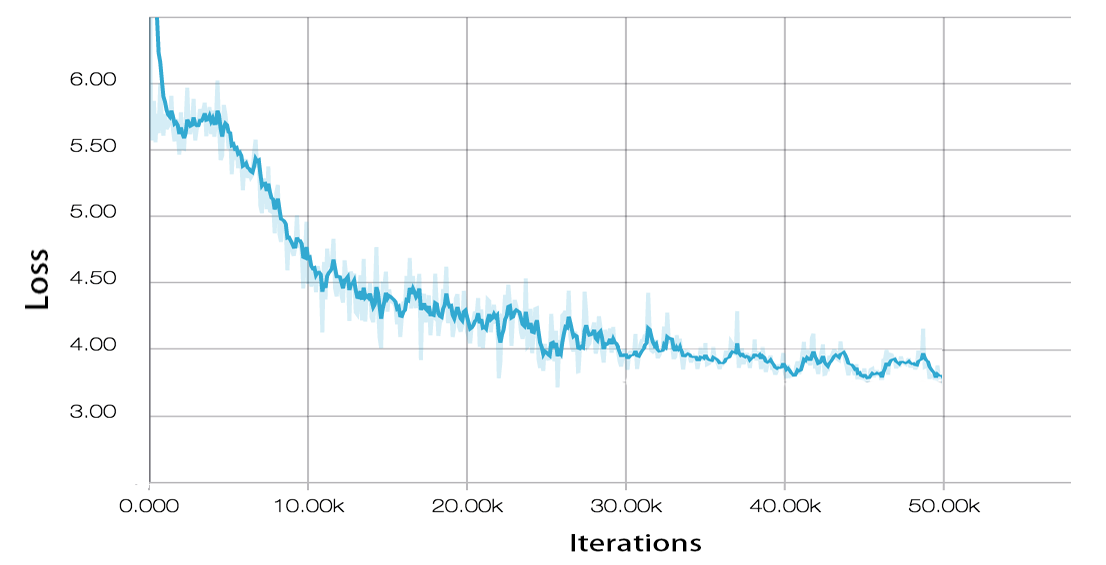}\\
  \caption{
    Convergence curve of our proposed DQN based localization.
  }
  \label{converge}
\end{figure}

\subsection{Implementation Details}
\textbf{Image Preprocessing} First, the images are resized to [224, 224] and fed into Localization stage driven by DQN, where interpolation are performed. After that, we convert the 12-bit CT images into 3-channel 8-bit input following \cite{zhou2017fixed}, which preserves most of abdominal organs. The pixel values in a CT slice were clipped to [-100, 240], then re-scaled to [0, 255], and finally duplicated 3 times to form 3 channels. Our framework is able to accept unfixed image size, so no more pre-processing is needed.

\textbf{Localization Network} With regards to DQN-driven localization, the region descriptor is generated with the backbone model VGG-16\cite{VGG} pretrained on ImageNet\cite{ImageNet,NIPS2012_4824}.  New layers of our three axis-specific Q-network are initialized by random weights with Gaussian distribution of 0 mean and 0.001 standard deviation. Each Q-network is trained for 25 epochs, and each epoch is ended after performing maximal 10 actions in each slice. During $\epsilon$-greedy training, $\epsilon$ decays from 1 to 0.1 with the a decrease rate of 0.1 per epoch in first 10 epochs, and the agent will keep on exploring with $\epsilon=0.1$ for the rest epochs. The discount factor $\gamma$ is set to 0.9. The replay memory size is set to 800,000, which contains about 1 epoch of transitions. The mini batch size in training is set to 100. For the trigger action, we consider a threshold $\tau_{IoU}$ = 0.2 for modified IoU and $\tau_{Recall}$ = 0.9 for Recall rate as positive localization.

\textbf{Segmentation Network} After the DQN-driven localization phase, CT slices are cropped and fed into the deformable U-Net. During training phase, in order to lose as little information as possible, we multiply the width and height of the bounding box computed by localization phase by a factor of $1.5$. We train the deformable U-Net with 2 optimizers, Momentum and Adam \cite{kingma2014adam}. The momentum optimizer gets a better result with learning rate 0.001, exponential decay rate 0.95 and momentum 0.9, training for 50 epochs. Both DQN and deformable U-Net are implemented in Tensorflow framework \cite{tensorflow}. Hyperparameters were determined with grid search. An early-stopping scheme was also adopted, which means we stop training when the performance on validation set has stopped improving. All four folds of the cross-validation were trained with the same network architecture, hyper-parameters and stopping criteria.

\subsection{Localization } 

\begin{figure*}[ht]
\centering
\begin{minipage}[t]{0.22\textwidth}
  \centering
  \includegraphics[width=3cm]{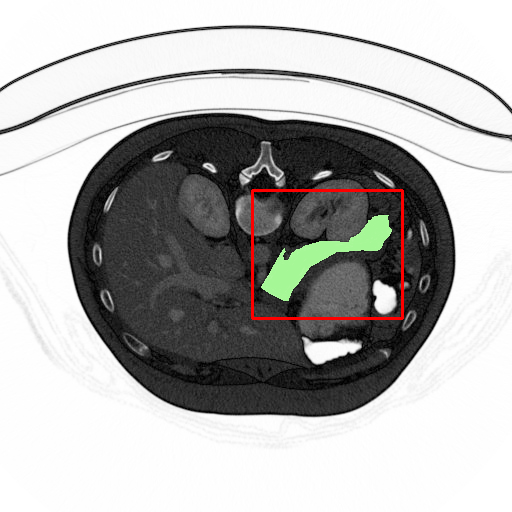}
  \end{minipage}
\begin{minipage}[t]{0.22\textwidth}
  \centering
  \includegraphics[width=3cm]{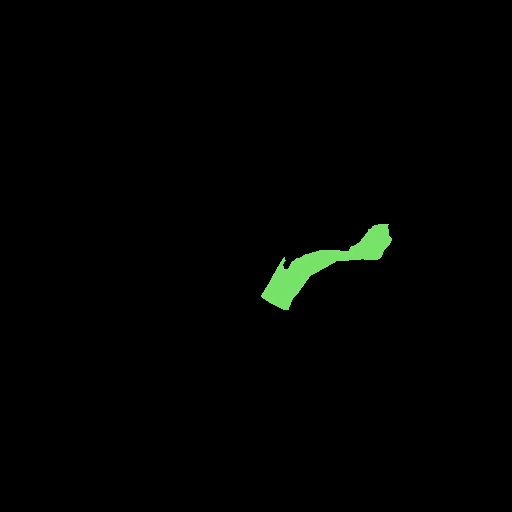}
  \end{minipage}
\begin{minipage}[t]{0.22\textwidth}
  \centering
  \includegraphics[width=3cm]{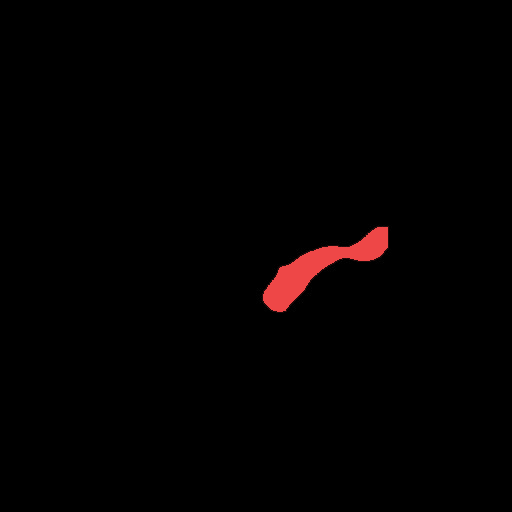}
  \end{minipage}
\begin{minipage}[t]{0.22\textwidth}
  \centering
  \includegraphics[width=3cm]{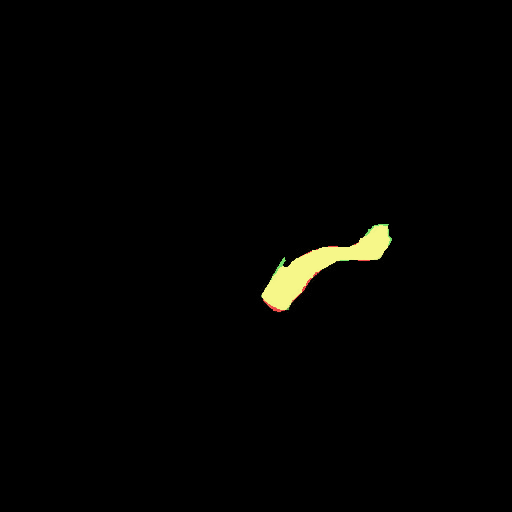}
  \end{minipage}
\begin{minipage}[t]{0.22\textwidth}
  \centering
  \includegraphics[width=3cm]{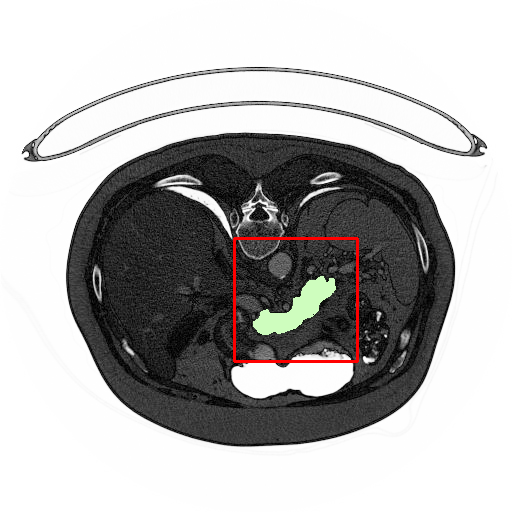}
  \caption*{Original}
  \end{minipage}
\begin{minipage}[t]{0.22\textwidth}
  \centering
  \includegraphics[width=3cm]{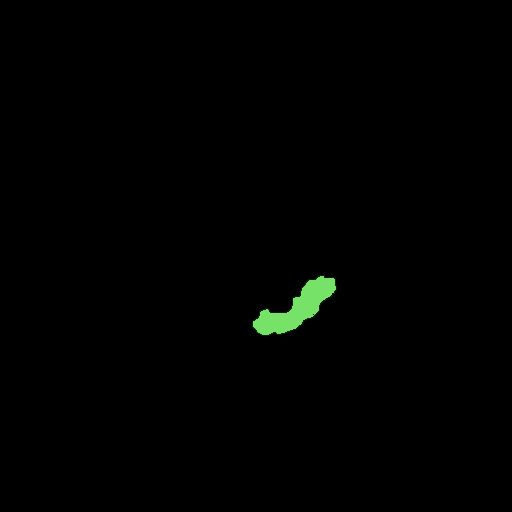}
  \caption*{Ground truth}
  \end{minipage}
\begin{minipage}[t]{0.22\textwidth}
  \centering
  \includegraphics[width=3cm]{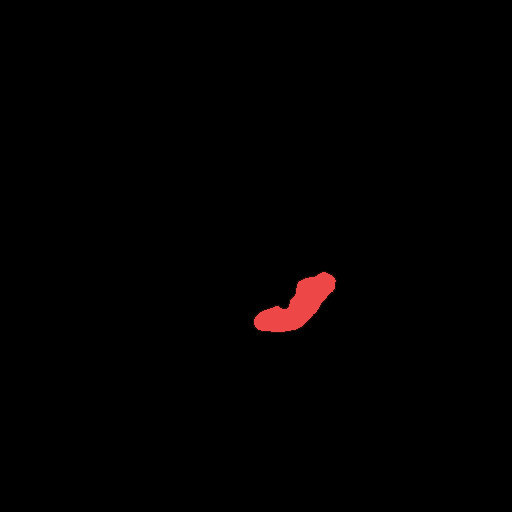}
  \caption*{Prediction map}
  \end{minipage}
\begin{minipage}[t]{0.22\textwidth}
  \centering
  \includegraphics[width=3cm]{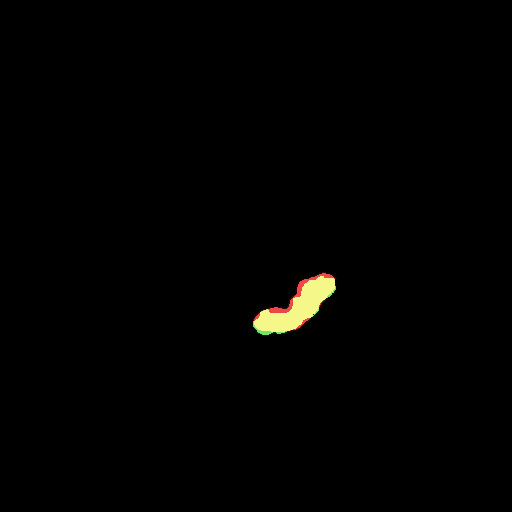}
  \caption*{Overlayed region}
  \end{minipage}
\caption{2D Illustration of segmentation result (best viewed in color). Two rows correspond to two different slices of pancreas. First column. Original input CT image, with pancreas area marked green, and bounding box given by localization process marked red. Second column. Ground truth pancreas map. Third column. Prediction map generated by our model. Last column. Overlayed prediction map and ground truth, with overlapped region marked yellow. The DSC reaches 94.32\% for the slice on the first raw, and 88.67\% for the second.}
\label{Seg_Visual}
\end{figure*}

\textbf{Model convergence.}
The MDP formulation given in section 3.2 reasonably simplified the agent's environment, which lowers the probability of unstable learning. Besides, the locations of pancreas is highly similar among different CT scans. This prior generates a direction preference in localization sequences, which helps limit the agent's exploration space thus guarantees convergence. We also adopted $\epsilon$-greedy and replay memory techniques to achieve stable learning. $\epsilon$-greedy with epoch-decayed exploration allows for more experience accumulation at early training stages. Replay memory lowers the oscillation caused by correlations between consecutive experiences. Fig.\ref{converge} indicates a good convergence.

\textbf{Localization sequence and action correlation.} Fig.\ref{sequence} shows an axial sample testing sequence. The sequence ends with a tight bounding box of the target, and also shows a combination of various actions. We hereby further analyze the correlation of different actions by taking the axial data as an example. We first calculate the action frequency vector of each localization sequence, and then derives the Pearson correlation metric to capture significant associations between different action pairs. High correlations between action pairs like zoom down-right and shift left are visualized in Fig.\ref{freq}. As we assumed in Sec. 3.2, the agent tends to combine zoom and shift actions of opposite directions (e.g, zoom down-right and shift left) to better zoom to the pancreas. Precise localization is obtained through the game between this kind of opposite-direction action pairs.

\textbf{Localization for different sizes of pancreas.} To illustrate the performance of our DQN-localization agent, we especially show the localization results of Volume 76's axial slices in Fig.\ref{volume76}. With Volume 76's axial slices, we achieve an average recall rate of 89.7\% and an average IoU of 0.11, which indicates that the main part of our target pancreas is retrieved. Note that axial scenario is the hardest one among three axes, however, sample results show stable bounding boxes despite the variance in pancreas size. The different sizes of triggered bounding boxes suggest that our agent may terminate without exhausting all the steps, and has learned the proper time to initiate a trigger action. Quantitative results of all testing samples are further discussed in Sec. 4.5.

\textbf{Modest and controllable randomness.} RL-like algorithms often suffer from a large search space and randomness, which may affect the model's performance. However, as indicated by our experiments, the randomness in our case is modest and controllable. Although a small amount of randomness is introduced because of $\epsilon$-greedy strategy, it decays as the training proceeds, namely the randomness is gradually reduced and replaced by the learned strategy. Thus, our model will finally converge at a stable state. Our experiments also support this claim. We used 4-fold cross-validation to test whether the algorithm is stable and well-trained. Based on the good testing results achieved in multiple times' training, we verify that our DQN model is stable and the performance is reproducible.

\subsection{Segmentation Results}
The final result is shown in Table \ref{result_table}. After 50 epochs' training, the mean DSC result of our model has reached $86.93 \pm 4.92$, which outperforms the state-of-the-art 3D coarse to fine model \cite{zhu20173d}. With tight and accurate bounding boxes given by localization stage, mean DSC increases by 2.34\%, and with the deformable structure, the worst case is able to perform better than ever. Note that the variance is relatively large. This is caused by the mixture of hard and easy cases. However for the evaluation of the same image, our method generally gets a higher DSC than method specified in [3][4]. The 2D illustration of segmentation result is shown in Fig \ref{Seg_Visual}.

\begin{table}[!t]
    \setlength{\belowdisplayskip}{-0.1cm}
    \scriptsize
    \caption{Evaluation results (measured by DSC, \%) of pancreas segmentation on NIH dataset. Our result is compared with recent state-of-the-art pancreas segmentation methods, and shows the greatest performance.}
    \centering
    \begin{tabular}{p{0.19\textwidth}p{0.06\textwidth}<{\centering}p{0.06\textwidth}<{\centering}p{0.085\textwidth}<{\centering}}
        \hline
        Method & Min DSC & Max DSC & \textbf{Mean DSC} \tabularnewline
        \hline\hline
        Roth \ul{et al.}, MICCAI'2015\cite{roth2015deeporgan} & 23.99 & 86.29 & $71.42 \pm 10.11$ \tabularnewline
        \hline
        Roth \ul{et al.}, MICCAI'2016\cite{roth2016spatial} & 34.11 & 88.65 & $78.01 \pm 8.20$ \tabularnewline
        \hline
        Dou \ul{et al.}, MIC'2017\cite{dou20163d} & 62.53 & 90.32 & $82.25 \pm 5.91$ \tabularnewline
        \hline
        Zhou \ul{et al.}, MICCAI'2017\cite{zhou2017fixed} & 62.43 & 90.85 & $82.37 \pm 5.68$ \tabularnewline
        \hline
        Zhu \ul{et al.}, Arxiv'2017\cite{zhu20173d} & 69.62 & \textbf{91.45} & $84.59 \pm 4.86$ \tabularnewline
        \hline\hline
        \textbf{Ours} & \textbf{74.32} & 91.34 & \textbf{86.93 $\pm$ 4.92} \tabularnewline
        \hline
    \end{tabular}
    \label{result_table}
\end{table}

\subsection{Ablation test}
In order to justify the coordination between two stages of our framework, we conduct ablation experiments on separate stages.

\textbf{Test on localization backbone} In order to test the localization backbone, the DQN localization model is replaced with Faster-RCNN. Rather than training Faster-RCNN end-to-end from scratch, we initialize the model with weights obtained from pretraining on the ImageNet dataset \cite{ImageNet} and then fine-tune the network with our NIH dataset. We set the anchor boxes in the RPN to be 256, and train for 50 epochs using stochastic gradient descent with a learning rate of 0.0001 and momentum of 0.9. We tuned several hyperparameters including the number of ROI per image and weight decay parameters. As shown in Table \ref{recall_table}, the performance measured by Recall all declines, especially for the toughest axial scenario. With axial slices, the Mean Recall suffers a drop from 78.11\% to 71.21\%, and the worst case shows a Recall of only 29.26\%, losing most of the pancreas. Agent for the coronal data shows the best localization performance for both two methods, mainly because the pancreas occupies a higher fraction of the slice than other two axes. This experiment proves the capability of our DRL agent in geometry-aware localization of pancreas with high anatomical variability. For a target object with similar positions in different testing images, a top-down DRL localization sequence is more robust than a Mask-RCNN method dirven by a bottom-up Region Proposal Network.

\begin{table}[!t]
    \scriptsize
    \caption{Evaluation of localization results of three axes with DQN-driven agent and Faster-RCNN. Performance is measured by Recall, and the DQN-driven agent shows better results in all scenarios.}
    \centering
    \begin{tabular}{ p{0.08\textwidth} p{0.05\textwidth} p{0.075\textwidth}<{\centering} p{0.075\textwidth}<{\centering} p{0.1\textwidth}<{\centering}}
        \hline
        Method & View & Min Recall & Max Recall & \textbf{Mean Recall} \tabularnewline
        \hline\hline
        DQN(Ours) & Axial &\textbf{68.32} & 91.79 & \textbf{78.11 $\pm$ 9.4} \tabularnewline
        \hline
        DQN(Ours)& Sagittal &\textbf{72.91} & \textbf{93.21} & \textbf{82.69 $\pm$ 8.42} \tabularnewline
        \hline
        DQN(Ours)& Coronal &\textbf{77.99} & \textbf{94.43} & \textbf{86.91 $\pm$ 4.85} \tabularnewline
        \hline\hline

        Faster-RCNN & Axial & 29.26 & \textbf{93.75} &  71.21 $\pm$ 16.75 \tabularnewline
        \hline
        Faster-RCNN & Sagittal & 38.67 & 88.53 & 74.54 $\pm$ 10.03  \tabularnewline
        \hline
        Faster-RCNN & Coronal & 33.58 & 93.98 & 79.09 $\pm$ 11.44 \tabularnewline 
        \hline

    \end{tabular}
    \label{recall_table}
\end{table}

\begin{table}[!t]
    \caption{Ablation studies of two stages.}
    \scriptsize
    \centering
    \begin{tabular}{ p{0.18\textwidth} p{0.058\textwidth}<{\centering} p{0.058\textwidth}<{\centering} p{0.09\textwidth}<{\centering}}
        \hline
        Methods & Min DSC & Max DSC& \textbf{Mean DSC} \tabularnewline
        \hline\hline
        Mask-RCNN & 63.27 & 82.76 & 75.81 $\pm$ 4.90  \tabularnewline
        Mask-RCNN + deformable & 64.25 & 90.84 & 76.69 $\pm$ 5.70  \tabularnewline
        DRL + non-deformable & 68.91 & 91.32 & 85.43 $\pm$ 5.92 \tabularnewline
        \hline
        DRL + deformable (Ours) & \textbf{74.32} & \textbf{91.34} & \textbf{86.93 $\pm$ 4.92} \tabularnewline
        \hline
    \end{tabular}
    \label{ablation_table}
\end{table} 

Experiments are also done by removing localization phase, as expected, the model cannot converge to the right direction due to the high class imbalance. This indicate that the main bottleneck of this problem lies in the first localization stage. The better the bounding boxes are, the better the segmentation result will be.

\textbf{Test on end-to-end model} We validate the effectiveness and performance of our model by separately replacing the localization stage and segmentation stage with their alternatives. This ablation study tests the relationship and relative significance of two stages. As shown in Table \ref{ablation_table}, if we fix Segmentation part and replace DRL with Mask-RCNN, the DSC drops significantly (Mean DSC 10\%). If we fix DRL part and replace deformable part, there is also a drop on DSC, however not so severe (Mean DSC 1.5\%). If we remove both the DRL localization and deformable convolution, the Max DSC drops nearly 10\%, which proves the effectiveness of the proposed method. The results also indicate that two stages of our model are highly complementary. 

We also indicate that a modest amount of randomness in the localization stage won't harm the final segmentation result. As indicated by Table \ref{result_table} and Table \ref{recall_table}, the fine segmentation stage can achieve a promising result as long as the coarse localization has a high recall rate of the pancreas, so modest randomness in the bounding boxes’ location will not harm the final segmentation.

\section{Discussion and Conclusions}
This paper proposes an innovative deep Q learning(DQN) driven and deformable U-Net based approach for pancreas segmentation. The main motivation behind the model is to design a context-adaptive, environment-interactive and robust pipeline to address the challenges in pancreas segmentation. The DQN driven localization agent learns to dynamically and adaptively adjust the bounding box, and deformable U-Net based segmentation extracts the geometry-aware features from the bounding box given by DQN. Experiments show that the proposed model gives state-of-the-art pancreas segmentation result and especially improves the result on hard tasks. Ablation experiments further validate the effectiveness of our model on interaction with contextual pancreas-related environment and extraction of anisotropy features.

The proposed method shows high sensitivity and specificity in localization, and also high overlap in segmentation between the automatically generated results and the manual annotations. As shown in Table \ref{result_table}, even though \cite{zhu20173d} has a better performance than ours in the easiest cases, we show significant improvement in the segmentation of those hard ones with smaller pancreas size and higher morphological irregularity. Besides, we believe there is space for exploration in formulating the object localization task into Markov Decision Process, as well as the way to represent the prior of the target's position. Future work will also investigate whether integrating the localization and segmentation stages into an end-to-end framework might be beneficial.

\section{Acknowledgement}
This work is supported in part by: Zhejiang Provincial Natural Science Foundation of China under Grant LR19F020004, the National Basic Research Program of China under Grant 2015CB352302, the National Natural Science Foundation of China under Grants (U1509206 and 61751209), Zhejiang University K.P.Chao's High Technology Development Foundation, and Tencent AI Lab Rhino-Bird Joint Research Program(No. JR201806). 
The corresponding author of this paper is Xi Li.

\bibliographystyle{IEEEtran}  
\bibliography{sample}

\end{document}